\documentclass[11pt,a4paper]{article}
\usepackage[utf8]{inputenc}
\usepackage[T1]{fontenc}
\usepackage{amsmath}
\usepackage{amssymb}
\usepackage{graphicx}
\usepackage{booktabs}
\usepackage{array}
\usepackage{tabularx}
\usepackage{caption}
\usepackage{geometry}
\usepackage{enumitem}
\usepackage{hyperref}
\title{SKINOPATHY AI: Smartphone-Based Ophthalmic Screening and Longitudinal Tracking Using Lightweight Computer Vision}
\author{
S. Kalaycioglu$^{1,2,3,4}$, C. Hong$^{5,6}$, M. Zhu$^4$, H. Xie$^4$ \\
$^1$Toronto Metropolitan University \\
$^2$York University \\
$^3$AIMechatroniX Inc. \\
$^4$DrRobot Inc \\
$^5$Skinopathy Inc. \\
$^6$Scarborough Health Network, Centenary Hospital, Scarborough, Ontario, Canada
}
\date{}

\begin{document}

\maketitle

\begin{abstract}
Early ophthalmic screening in low-resource and remote settings is constrained by access to specialized equipment and trained practitioners. We present SKINOPATHY AI, a smartphone-first web application that delivers five complementary, explainable screening modules entirely through commodity mobile hardware: (1) redness quantification via LAB a* color-space normalization; (2) blink-rate estimation using MediaPipe FaceMesh Eye Aspect Ratio (EAR) with adaptive thresholding; (3) pupil light reflex characterization through Pupil-to-Iris Ratio (PIR) time-series analysis; (4) scleral color indexing for icterus and anemia proxies via LAB/HSV statistics; and (5) iris-landmark-calibrated lesion encroachment measurement with millimeter-scale estimates and longitudinal trend tracking. The system is implemented as a React/FastAPI stack with OpenCV and MediaPipe, MongoDB-backed session persistence, and PDF report generation. All algorithms are fully deterministic, privacy-preserving, and designed for non-diagnostic consumer triage. We detail system architecture, algorithm design, evaluation methodology, clinical context, and ethical boundaries of the platform. SKINOPATHY AI demonstrates that multi-signal ophthalmic screening is feasible on unmodified smartphones without cloud-based AI inference, providing a foundation for future clinically validated mobile ophthalmoscopy tools.
\end{abstract}

\textbf{Index Terms} --- mobile ophthalmology, computer vision, MediaPipe, OpenCV, blink rate, pupil reflex, scleral icterus, pterygium, tele-triage, mHealth, explainable AI, lesion tracking

\section{INTRODUCTION}
Globally, an estimated 2.2 billion people live with vision impairment or blindness, the majority of whom reside in low- and middle-income countries with limited access to qualified ophthalmologists [1]. Even in technologically advanced nations, barriers including geographic distance, cost, and shortage of specialists result in delayed diagnosis of conditions that are both common and treatable when identified early. Conditions such as dry eye disease, conjunctivitis, pterygium, and jaundice-associated scleral icterus all carry objective, measurable ocular manifestations that can, in principle, be estimated from a photograph or short video captured by a consumer smartphone.

Existing mobile ophthalmology tools typically fall into one of two categories: hardware attachments that convert a smartphone into a slit-lamp or fundus camera, or cloud-based AI services that require uploading sensitive biometric imagery. Both approaches introduce barriers to adoption. Hardware attachments impose cost and availability constraints. Cloud services introduce latency, privacy risks, and dependence on network connectivity. There is a compelling gap for a lightweight, privacy-preserving, explainable screening application that operates entirely within the browser and performs all inference locally on the device or on a minimal self-hosted backend.

SKINOPATHY AI addresses this gap by delivering a five-module ophthalmic screening pipeline implemented using freely available computer vision libraries (OpenCV, MediaPipe) on standard React/FastAPI infrastructure. The application is explicitly positioned as a consumer screening and self-triage tool --- not a diagnostic device --- and implements conservative triage messaging, explicit consent gating, and non-identifiable session storage. Every metric produced by the system is derived from an interpretable, documented algorithm rather than an opaque neural network, enabling clinicians and researchers to audit, validate, and extend the methodology.

The primary contributions of this work are:
\begin{itemize}[leftmargin=*]
\item A multi-module, explainable ophthalmic screening pipeline deployable on commodity smartphones without specialized hardware or cloud AI inference;
\item An iris-landmark-calibrated lesion encroachment estimation algorithm producing millimeter-scale measurements from a single photograph;
\item An adaptive, fps-aware blink detection algorithm based on EAR with temporal smoothing;
\item A longitudinal session architecture enabling trend detection for lesion progression across clinical visits;
\item A privacy-first, consent-gated architecture suitable for consumer health applications;
\item An evaluation framework and ablation design suitable for clinical validation studies.
\end{itemize}

The remainder of this paper is organized as follows. Section 2 surveys related work. Section 3 describes system architecture and data flow. Section 4 details the algorithms of each screening module. Section 5 presents the experimental design and evaluation methodology. Section 6 reports results and analysis. Section 7 discusses clinical relevance, limitations, and future directions. Section 8 addresses ethical and privacy considerations. Section 9 concludes the paper.

\section{RELATED WORK}
\subsection{Mobile Ophthalmology}
The use of smartphones for ophthalmic screening has expanded significantly over the past decade. Bolster et al. [2] demonstrated that smartphones equipped with lens attachments can achieve fundus imaging quality sufficient for diabetic retinopathy screening. More recent work from Raju et al. [3] and colleagues has extended this approach to slit-lamp photography using custom attachment optics. While these hardware-assisted approaches yield high image quality, their dependency on physical accessories limits deployment in resource-constrained environments.

Software-only approaches have also attracted significant research attention. Work by Toslak et al. [4] explored peripheral retinal imaging using standard smartphone optics with no attachment. In the anterior segment, several studies have demonstrated the feasibility of using smartphone photographs for conjunctival redness grading [5], pterygium staging [6], and scleral icterus estimation [7]. SKINOPATHY AI synthesizes and extends this body of work by integrating multiple anterior-segment screening signals into a single, session-aware application.

\subsection{Eye Aspect Ratio and Blink Detection}
The Eye Aspect Ratio (EAR) was introduced by Soukupova and Cech [8] as a reliable, landmark-based blink detector. EAR is computed from six periorbital landmarks identified by facial landmark detectors, and a blink is registered when EAR falls below a threshold for a sufficient number of consecutive frames. Subsequent work has refined threshold selection strategies, including adaptive and percentile-based methods [9]. MediaPipe FaceMesh [10], developed at Google, provides a 468-landmark facial mesh that enables real-time, accurate periorbital landmark tracking on mobile devices, making EAR-based blink detection practical in browser-based applications.

\subsection{Pupil Light Reflex Analysis}
The pupillary light reflex (PLR) is a well-characterized neurological indicator with clinical relevance in head injury assessment, pharmacological monitoring, and autonomic nervous system evaluation [11]. Portable pupillometry devices such as the NPi-300 (NeurOptics) represent the clinical gold standard. More recently, video-based methods using standard cameras have shown promise for PLR estimation [12][13]. The Pupil-to-Iris Ratio (PIR) approach used in SKINOPATHY AI is adapted from these methods, leveraging MediaPipe iris landmark refinement for robust iris diameter estimation without additional calibration hardware.

\subsection{Scleral Color Analysis}
Quantitative scleral color analysis for icterus screening has been explored as a low-cost alternative to serum bilirubin measurement [7][14]. Methods typically involve isolating the scleral region, converting to a perceptually uniform color space (such as CIE LAB), and computing statistics on the b* channel (blue-yellow axis). Pallor estimation from conjunctival or scleral appearance has similarly been proposed as a proxy for anemia [15]. SKINOPATHY AI implements both indices with explicit disclaimer messaging regarding the confounding effects of device white balance, ambient lighting, and individual variation.

\subsection{Pterygium and Lesion Measurement}
Pterygium is a fibrovascular overgrowth of the bulbar conjunctiva onto the corneal surface, with prevalence strongly correlated with ultraviolet light exposure [16]. Clinical grading systems including the Johnston and Tan scales rely on measurement of corneal encroachment distance relative to the limbus. Several image-based measurement approaches have been proposed [17][18], typically requiring calibration from a known reference or slit-lamp imaging. SKINOPATHY AI introduces an iris-diameter-based pixel-to-millimeter calibration derived from the well-characterized horizontal visible iris diameter (HVID) distribution, enabling single-image encroachment estimation without additional hardware.

\section{SYSTEM ARCHITECTURE}
\subsection{Overview}
SKINOPATHY AI follows a three-tier architecture: a React frontend for user interaction and media capture, a Python/FastAPI backend for all image and video processing, and a MongoDB database for session and result persistence. Figure 1 provides a schematic overview. All communication between frontend and backend occurs over a REST API with routes prefixed under /api, enabling transparent Kubernetes ingress path-based routing. The frontend communicates with the backend exclusively through the REACT\_APP\_BACKEND\_URL environment variable, ensuring no hardcoded URLs or ports in application code.

\begin{table}[htbp]
\caption{System Technology Stack by Tier}
\label{tab:techstack}
\centering
\begin{tabularx}{\textwidth}{l X X}
\toprule
Tier & Technology & Role \\
\midrule
Frontend & React, Chart.js, MediaRecorder API, Axios & UI, consent, media capture, visualization \\
Backend & FastAPI, OpenCV, MediaPipe FaceMesh, ReportLab & Image/video processing, metrics, PDF export \\
Database & MongoDB (Motor async client), UUID keys & Session storage, result persistence, trend data \\
DevOps & Supervisor, Kubernetes ingress, .env variables & Process management, routing, configuration \\
\bottomrule
\end{tabularx}
\end{table}

\begin{figure}[htbp]
\centering
\includegraphics[width=1.05\textwidth]{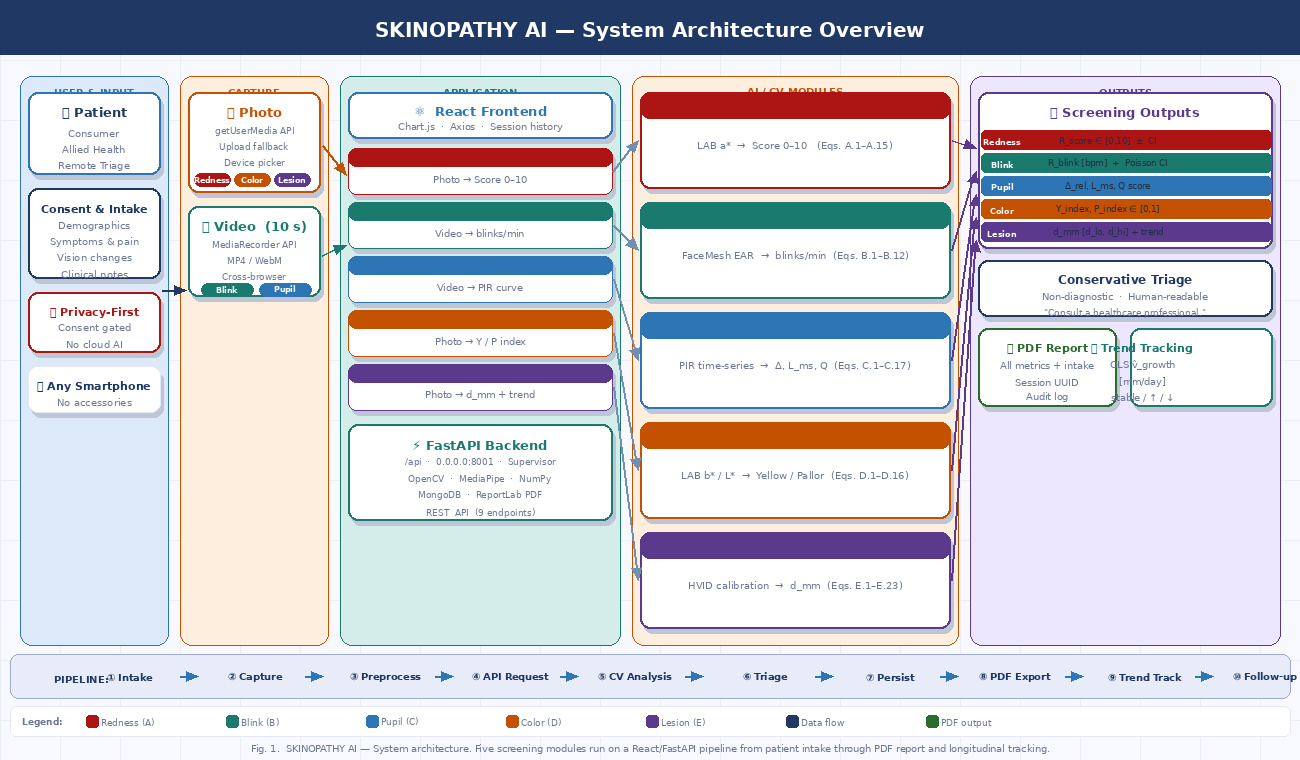}
\caption{Schematic overview of SKINOPATHY AI three-tier architecture.}
\label{fig:arch}
\end{figure}

\subsection{Frontend}
The React frontend provides five module views corresponding to the five screening functions, plus a consent and symptom intake form, session history browser, and PDF report download. Camera access is managed through the browser-native getUserMedia API. Video recording uses the MediaRecorder API with cross-browser fallback handling. A global stream manager prevents concurrent camera stream conflicts when switching between modules. Device pickers allow users to select from available cameras (front/rear/external), which is important for reliable rear-camera capture of the eye from close range.

Metric visualizations use react-chartjs-2 with Chart.js for the PIR time-series chart and trend summaries. All API calls are routed through a centralized Axios configuration that reads REACT\_APP\_BACKEND\_URL at build time. UI supports light and dark themes, with preference persisted in localStorage.

\subsection{Backend}
The FastAPI backend exposes all routes under the /api prefix and binds to 0.0.0.0:8001 under Supervisor process management. Endpoints are organized into two groups: session management endpoints (POST /api/sessions, GET /api/sessions, GET /api/sessions/\{id\}, GET /api/sessions/\{id\}/report.pdf) and analysis endpoints (POST /api/analyze/redness, /api/analyze/blink, /api/analyze/pupil, /api/analyze/color, /api/analyze/lesion). All analysis endpoints accept multipart form uploads of image or video files along with a session\_id parameter.

Image processing uses OpenCV (cv2) for color space conversion, morphological operations, and region-of-interest extraction. MediaPipe FaceMesh with refine\_landmarks=True is used for landmark detection across all video-based modules and for iris detection in the lesion module. PDF generation uses ReportLab, with each section populated from the session document retrieved from MongoDB.

\subsection{Data Flow}
The end-to-end data flow through SKINOPATHY AI proceeds as follows: (1) the user completes the consent and intake form, creating a new session document in MongoDB with a UUID primary key and ISO 8601 timestamp; (2) for each screening module, the user performs the guided capture (photo upload or 10-second video recording); (3) the captured media is posted to the corresponding /api/analyze endpoint along with the session\_id; (4) the backend performs preprocessing, runs the module algorithm, and returns metrics in a JSON payload; (5) the frontend displays results and saves them to the session document; (6) the user may download a PDF report that aggregates all session results with intake data and triage guidance.

\section{SCREENING MODULE ALGORITHMS}
\subsection{Notation and Common Definitions}
The following notation is used uniformly across all five module formulations. All pixel coordinates are in image space with origin at the top-left corner. Color space transforms are performed by OpenCV with 8-bit unsigned integer representations unless otherwise stated.

\begin{table}[htbp]
\caption{Unified Notation for All Module Algorithms}
\label{tab:notation}
\centering
\begin{tabular}{p{2.5cm} p{4.5cm} p{8cm}}
\toprule
Symbol & Domain / Units & Definition \\
\midrule
I & $H\times W\times 3$, uint8 & Input BGR image from smartphone camera \\
I$_{\text{LAB}}$ & $H\times W\times 3$, float & CIE LAB image; OpenCV scaling: $L \in [0,255]$, $a^* \in [0,255]$, $b^* \in [0,255]$ \\
I$_{\text{HSV}}$ & $H\times W\times 3$, uint8 & HSV image; $H \in [0,179]$, $S \in [0,255]$, $V \in [0,255]$ \\
M & $H\times W$, bool & Binary mask; M(x,y)=1 if pixel belongs to region of interest \\
$\mathbb{E}_M[\cdot]$ & --- & Spatial expectation over mask: $(1/N_M) \sum_{(x,y): M(x,y)=1} f(x,y)$ \\
p$_i$ & $\mathbb{R}^2$ & Unnormalized pixel-space landmark coordinate (x$_i$, y$_i$) \\
$\|\cdot\|_2$ & --- & Euclidean distance between two 2D points \\
clip(v,a,b) & --- & Saturating clamp: max(a, min(b, v)) \\
fps & frames/s & Estimated frame rate of the captured video \\
t$_k$ & seconds & Timestamp of frame k: t$_k$ = k / fps \\
\bottomrule
\end{tabular}
\end{table}

\subsection{Redness Quantification (Hyperemia Screening)}
\subsubsection{Color Space Selection and sRGB$\rightarrow$LAB Derivation}
Conjunctival hyperemia produces a spectral reflectance shift towards 550--700 nm. The CIE 1976 L*a*b* color space is selected because the a* axis encodes the red-green opponent channel independently of luminance L*, decoupling chromatic redness from lighting variation. OpenCV's cv2.COLOR\_BGR2LAB applies a two-stage transform. For channel c with raw value $v_c \in [0,255]$, the linearized sRGB value is:

\[ \hat{v}_c = \frac{v_c}{255} \quad (1) \]

\[ v_{\text{lin},c} = \begin{cases}
\hat{v}_c / 12.92, & \text{if } \hat{v}_c \leq 0.04045 \quad (2) \\
\left( (\hat{v}_c + 0.055) / 1.055 \right)^{2.4}, & \text{otherwise} \quad (3)
\end{cases} \]

The linear sRGB triplet is projected to CIE XYZ under D65 illuminant via the standard matrix:

\[ \begin{pmatrix} X \\ Y \\ Z \end{pmatrix} = 
\begin{pmatrix} 
0.4124 & 0.3576 & 0.1805 \\ 
0.2126 & 0.7152 & 0.0722 \\ 
0.0193 & 0.1192 & 0.9505 
\end{pmatrix}
\begin{pmatrix} R_{\text{lin}} \\ G_{\text{lin}} \\ B_{\text{lin}} \end{pmatrix} \quad (4) \]

Normalized XYZ values are passed through the CIE f function and the LAB axes are derived as:

\[ f(t) = \begin{cases}
t^{1/3} & \text{if } t > (6/29)^3 \\
(1/3)(29/6)^2 \cdot t + 4/29 & \text{otherwise}
\end{cases} \quad (5) \]

\[ L^* = 116 \cdot f(Y/Y_n) - 16 \quad [Y_n = 1.0] \quad (6) \]

\[ a^* = 500 \cdot [f(X/X_n) - f(Y/Y_n)] \quad [X_n = 0.9505] \quad (7) \]

\[ b^* = 200 \cdot [f(Y/Y_n) - f(Z/Z_n)] \quad [Z_n = 1.0890] \quad (8) \]

OpenCV rescales to uint8 storage: L$_{\text{cv}}$ = $(L^*/100)\cdot 255$, a*$_{\text{cv}}$ = a* + 128, b*$_{\text{cv}}$ = b* + 128. Neutral gray maps to (128, 128, 128).

\subsubsection{Luminance-Weighted Scleral Mask}
A luminance-gated scleral mask isolates the conjunctival surface, excluding iris, pupil, and eyelids. Otsu's global method applied to the L* channel provides an adaptive threshold:

\[ \tau_L = \max(L_{\min}, \tau_{\text{Otsu}}(I_{\text{LAB}}[:,:,0])) \quad [L_{\min} = 160] \quad (9) \]

\[ M_{\text{sc}}(x,y) = 1 \quad \text{iff} \quad I_{\text{LAB}}(x,y,0) \geq \tau_L \quad \text{AND} \quad I_{\text{LAB}}(x,y,1) < 160 \quad (10) \]

The redness signal is computed as the luminance-weighted spatial mean of a* over M$_{\text{sc}}$, giving greater influence to brighter scleral pixels:

\[ w(x,y) = I_{\text{LAB}}(x,y,0) \cdot M_{\text{sc}}(x,y) \quad (11) \]

\[ \bar{a}^* = \frac{\sum_{x,y} w(x,y)\cdot I_{\text{LAB}}(x,y,1)}{\sum_{x,y} w(x,y)} \quad (12) \]

\[ \sigma_{a^*} = \sqrt{\mathbb{E}_M[(I_{\text{LAB}}(:,:,1) - \bar{a}^*)^2]} \quad (13) \]

\subsubsection{Score Normalization and Uncertainty Bounds}
The weighted mean a* is mapped to the [0, 10] redness score via linear normalization calibrated against Efron-graded images. Uncertainty bounds are propagated from $\sigma_{a^*}$:

\[ R_{\text{score}} = 10 \cdot \text{clip} \left( (\bar{a}^* - 120) / (150 - 120), 0, 1 \right) \quad (14) \]

\[ R_{\text{lo}} = 10 \cdot \text{clip} \left( (\bar{a}^* - \sigma_{a^*} - 120) / 30, 0, 1 \right) \quad (15) \]

\[ R_{\text{hi}} = 10 \cdot \text{clip} \left( (\bar{a}^* + \sigma_{a^*} - 120) / 30, 0, 1 \right) \quad (16) \]

Triage bands: 0--2 (normal), 2.1--4.0 (trace/mild --- monitor), 4.1--7.0 (moderate --- seek evaluation), 7.1--10.0 (severe --- seek prompt ophthalmic care).

\subsection{Blink Rate Estimation (Dry Eye Screening)}
\subsubsection{Eye Aspect Ratio: Vector Formulation}
MediaPipe FaceMesh provides 468 facial landmarks per frame. Six periorbital landmarks per eye define the EAR, where p$_1$ (outer canthus) and p$_4$ (inner canthus) span the horizontal fissure width, and p$_2$, p$_3$, p$_5$, p$_6$ are the upper and lower lid margins:

\[ \text{EAR}_k = \frac{\|p_{2,k} - p_{6,k}\|_2 + \|p_{3,k} - p_{5,k}\|_2}{2 \cdot \|p_{1,k} - p_{4,k}\|_2} \quad (17) \]

\[ \text{EAR}_k^{(\text{bilateral})} = (\text{EAR}_k^{(L)} + \text{EAR}_k^{(R)}) / 2 \quad (18) \]

For a fully open eye EAR $\approx 0.20$--$0.35$ (individual-dependent); during a complete blink EAR approaches 0. The bilateral mean is used to reduce unilateral occlusion artifacts.

\subsubsection{Temporal Smoothing and Adaptive Threshold}
A causal moving-average filter with fps-proportional window removes sub-frame jitter while preserving blink structure:

\[ \tilde{E}_k = \frac{1}{W_s} \cdot \sum_{j=k-W_s+1}^{k} \text{EAR}_j^{(\text{bilateral})} \quad (19) \]

\[ W_s = \max(1, \text{round}(0.04 \cdot \text{fps})) \quad (20) \]

The adaptive threshold is derived from a statistical baseline over the first N$_b$ frames (~1.5 s), using the median for robustness against blinks within the baseline window:

\[ \mu_E = \text{median}\{ \tilde{E}_k : k = 1, \dots, N_b \} \quad (21) \]

\[ \sigma_E = \sqrt{ \frac{1}{N_b-1} \sum_{k=1}^{N_b} (\tilde{E}_k - \mu_E)^2 } \quad (22) \]

\[ \tau_{\text{blink}} = \mu_E - \alpha \cdot \sigma_E \quad [\text{default } \alpha = 2.0] \quad (23) \]

\subsubsection{Finite-State Blink Detection and Rate Estimation}
A finite-state machine requires EAR to remain below the threshold for a minimum fps-adaptive frame count before registering a blink, preventing spurious detection from micro-movements or partial blinks:

\[ F_{\min} = \max(2, \text{round}(0.033 \cdot \text{fps})) \quad [\sim 33 \text{ ms min. blink duration}] \quad (24) \]

state$_k$ = BLINK iff $\tilde{E}_k < \tau_{\text{blink}}$ AND consecutive\_below $\ge F_{\min}$ (25)

Blink count B increments on each OPEN$\rightarrow$BLINK transition. The blink rate and its Poisson 95\% CI are:

\[ R_{\text{blink}} = (B / T) \cdot 60 \quad [\text{blinks/min}] \quad (26) \]

\[ \text{CI}_{\text{lo}} = (\chi^2_{0.025, 2B} / 2) \cdot (60/T) \quad (27) \]

\[ \text{CI}_{\text{hi}} = (\chi^2_{0.975, 2B+2} / 2) \cdot (60/T) \quad (28) \]

Risk strata: < 8 bpm (high dry-eye risk), 8--11 bpm (elevated), 12--20 bpm (normal), > 20 bpm (elevated --- irritation reflex).

\subsection{Pupil Light Reflex (Neurological Screening Signal)}
\subsubsection{PIR Definition and Normalization Rationale}
MediaPipe FaceMesh with refine\_landmarks=True yields one iris center c and four equidistant ring landmarks $\{r_1,\ldots,r_4\}$ per eye. The iris radius and eye width are:

\[ \rho_{\text{iris},k} = (1/4) \cdot \sum_{i=1}^{4} \|r_{i,k} - c_k\|_2 \quad (29) \]

\[ w_{\text{eye},k} = \|p_{\text{OC},k} - p_{\text{IC},k}\|_2 \quad [\text{outer to inner canthus}] \quad (30) \]

\[ \text{PIR}_k = (2 \cdot \rho_{\text{iris},k}) / w_{\text{eye},k} \in (0, 1) \quad (31) \]

Division by eye width makes PIR invariant to camera-subject distance changes within a session, and to head yaw within $\pm 15^\circ$ (beyond which perspective foreshortening introduces >5\% error in w$_{\text{eye}}$).

\subsubsection{Waveform Segmentation and PLR Metrics}
The baseline PIR and post-stimulus minimum are extracted as follows (nominal stimulus at t$_{\text{stim}}$ = 3 s):

\[ K_{\text{base}} = \{ k : t_k \in [3/\text{fps}, 1.5 \text{ s}] \} \quad (32) \]

\[ \text{PIR}_{\text{base}} = (1/|K_{\text{base}}|) \cdot \sum_{k \in K_{\text{base}}} \text{PIR}_k \quad (33) \]

\[ k^* = \arg\min_{k : t_k \geq 1.5 \text{ s}} \text{PIR}_k \quad (34) \]

\[ \text{PIR}_{\min} = \text{PIR}_{k^*} \quad (35) \]

PLR amplitude, latency, and constriction velocity:

\[ \Delta = \text{PIR}_{\text{base}} - \text{PIR}_{\min} \quad [\text{absolute constriction}] \quad (36) \]

\[ \Delta_{\text{rel}} = (\Delta / \text{PIR}_{\text{base}}) \cdot 100\% \quad [\% \text{ constriction}] \quad (37) \]

\[ L_{\text{ms}} = (t_{k^*} - t_{\text{stim}}) \cdot 1000 \quad [\text{latency, ms}] \quad (38) \]

\[ v_{\text{mean}} = \Delta / L_{\text{ms}} \quad [\text{mean velocity, PIR/ms}] \quad (39) \]

\[ v_{\max} = \max_{k \geq K_{\text{base}}} |(\text{PIR}_k - \text{PIR}_{k-1})\cdot\text{fps}| \quad [\text{peak velocity}] \quad (40) \]

\subsubsection{Quality Score and Optional Exponential Model}
A recording quality score $Q \in [0,1]$ gates unreliable results:

\[ Q_{\text{detect}} = (1/N_{\text{frames}}) \cdot \sum_k \mathbb{1}[\text{landmarks detected in frame } k] \quad (41) \]

\[ Q_{\text{stable}} = 1 - \text{clip}(\text{std}(w_{\text{eye},k})/\text{mean}(w_{\text{eye},k}), 0, 1) \quad (42) \]

\[ Q_{\text{resp}} = \mathbb{1}[\Delta_{\text{rel}} > 5\%] \quad (43) \]

\[ Q = (Q_{\text{detect}} + Q_{\text{stable}} + Q_{\text{resp}}) / 3 \quad (44) \]

For high-quality recordings ($Q \ge 0.8$), the constriction limb is fitted to an exponential decay model enabling stimulus-independent latency and time-constant estimation:

\[ \text{PIR}_{\text{fit}}(t) = \text{PIR}_{\min} + (\text{PIR}_{\text{base}}-\text{PIR}_{\min})\cdot\exp(-(t-t_{\text{stim}}-L_{\text{ms}}/1000)/\tau) \quad (45) \]

Parameters ($L_{\text{ms}}$, $\tau$) are estimated by non-linear least squares. This model is especially useful for pharmacological monitoring where precise $\tau$ comparison across sessions is required.

\subsection{Scleral Color Indices (Icterus and Anemia Proxies)}
\subsubsection{Scleral Mask via Three-Gate Intersection}
Accurate scleral color measurement requires excluding the iris, pupil, eyelid skin, and limbal vasculature. The scleral mask is the intersection of three spatial gates:

\[ M_{\text{bright}}(x,y) = \mathbb{1}[ I_{\text{LAB}}(x,y,0) \geq 190 ] \quad [\text{luminance gate}] \quad (46) \]

\[ M_{\text{low\_S}}(x,y) = \mathbb{1}[ I_{\text{HSV}}(x,y,1) \leq 60 ] \quad [\text{saturation gate}] \quad (47) \]

\[ M_{\text{sclera}} = \text{morphClose}(M_{\text{bright}} \cap M_{\text{low\_S}}, K_3) \quad [3\times3 \text{ close}] \quad (48) \]

\subsubsection{Icterus (Yellow) Index}
Hyperbilirubinemia deposits bilirubin in scleral connective tissue, shifting b* (blue-yellow axis) upward. The weighted mean and uncertainty are:

\[ \bar{b}^* = \mathbb{E}_{M_{\text{sclera}}}[ I_{\text{LAB}}(:,:,2) ] \quad (49) \]

\[ \sigma_{b^*} = \sqrt{ \mathbb{E}_{M_{\text{sclera}}}[ (I_{\text{LAB}}(:,:,2) - \bar{b}^*)^2 ] } \quad (50) \]

\[ Y_{\text{index}} = \text{clip} \left( (\bar{b}^* - 128) / (160 - 128), 0, 1 \right) \quad (51) \]

\[ Y_{\text{lo}} = \text{clip} \left( (\bar{b}^* - \sigma_{b^*} - 128) / 32, 0, 1 \right) \quad (52) \]

\[ Y_{\text{hi}} = \text{clip} \left( (\bar{b}^* + \sigma_{b^*} - 128) / 32, 0, 1 \right) \quad (53) \]

\subsubsection{Pallor Index and White Balance Correction}
Anemia-associated pallor produces reduced a* (less red) and elevated L* (paler). The composite pallor index combines both axes:

\[ \bar{L}^* = \mathbb{E}_{M_{\text{sclera}}}[ I_{\text{LAB}}(:,:,0) ] \quad (54) \]

\[ \bar{a}^* = \mathbb{E}_{M_{\text{sclera}}}[ I_{\text{LAB}}(:,:,1) ] \quad (55) \]

\[ L_{\text{term}} = \text{clip} \left( (\bar{L}^* - 200) / (240 - 200), 0, 1 \right) \quad (56) \]

\[ a_{\text{term}} = 1 - \text{clip} \left( (\bar{a}^* - 120) / (140 - 120), 0, 1 \right) \quad (57) \]

\[ P_{\text{index}} = 0.5\cdot L_{\text{term}} + 0.5\cdot a_{\text{term}} \quad (58) \]

A gray-world white balance correction is applied before LAB conversion to reduce inter-device bias. For channel $c \in \{B, G, R\}$:

\[ \bar{g}_{\text{ref}} = (\text{mean}_B + \text{mean}_G + \text{mean}_R) / 3 \quad (59) \]

\[ g_c = \bar{g}_{\text{ref}} / \bar{g}_c \quad (60) \]

\[ I_{\text{corr}}(x,y,c) = \text{clip}( I(x,y,c) \cdot g_c, 0, 255 ) \quad (61) \]

\subsection{Iris-Calibrated Lesion Encroachment and Tracking}
\subsubsection{Polar Coordinate System and Iris Calibration}
All spatial analysis in the lesion module operates in a polar coordinate system centered on the iris center landmark c = (c$_x$, c$_y$):

\[ \rho(x,y) = \|(x,y) - c\|_2 \quad [\text{pixels from iris center}] \quad (62) \]

\[ \theta(x,y) = \text{atan2}(y - c_y, x - c_x) \in [-\pi, \pi] \quad (63) \]

\[ \rho_{\text{iris}} = (1/4) \cdot \sum_{i=1}^{4} \|r_i - c\|_2 \quad (64) \]

The pixel-to-millimeter calibration factor uses the population mean HVID with full uncertainty propagation:

\[ \lambda = H_{\text{ref}} / (2\cdot\rho_{\text{iris}}) \quad [\text{mm/px}], \quad H_{\text{ref}} = 11.8 \text{ mm} \quad (65) \]

\[ \lambda_{\text{lo}} = 10.7 / (2\cdot\rho_{\text{iris}}) \quad [\text{mm/px, lower 95th CI bound}] \quad (66) \]

\[ \lambda_{\text{hi}} = 12.9 / (2\cdot\rho_{\text{iris}}) \quad [\text{mm/px, upper 95th CI bound}] \quad (67) \]

\[ \varepsilon_{\text{HVID}} = (\lambda_{\text{hi}} - \lambda_{\text{lo}}) / (2\cdot\lambda) \approx \pm9.3\% \quad [\text{relative uncertainty}] \quad (68) \]

When MediaPipe landmark detection fails, OpenCV HoughCircles is applied as a fallback, with an empirical +5\% upward correction on the estimated radius to compensate for eyelid occlusion of the iris boundary.

\subsubsection{Lesion Mask via Multi-Space Thresholding}
Pterygium/pinguecula tissue is characterized by high luminance, yellowish or pinkish hue, and low saturation relative to the conjunctival vasculature. The lesion mask is the intersection of three complementary color-space gates, refined by morphological operations:

\[ M_{\text{bright}}(x,y) = \mathbb{1}[ I_{\text{LAB}}(x,y,0) \geq 180 ] \quad (69) \]

\[ M_{\text{yellow}}(x,y) = \mathbb{1}[ I_{\text{LAB}}(x,y,2) \geq 140 ] \quad (70) \]

\[ M_{\text{red}}(x,y) = \mathbb{1}[ I_{\text{HSV}}(x,y,0) \in [0,25] \cup [155,179] ] \quad (71) \]

\[ M_{\text{low\_S}}(x,y) = \mathbb{1}[ I_{\text{HSV}}(x,y,1) \leq 80 ] \quad (72) \]

\[ M_{\text{raw}} = M_{\text{bright}} \cap (M_{\text{yellow}} \cup M_{\text{red}}) \cap M_{\text{low\_S}} \quad (73) \]

\[ M_{\text{lesion}} = \text{erode}(\text{dilate}(M_{\text{raw}}, K_3), K_3) ; \text{close}(\cdot, K_5) \quad (74) \]

\subsubsection{Near-Limbus Annular Analysis Band and Encroachment}
Encroachment is measured within the annular analysis band spanning 65--98\% of the iris radius from center, restricted to the nasal and temporal sectors where pterygium most commonly originates:

\[ \rho_{\text{lo}} = 0.65\cdot\rho_{\text{iris}}, \quad \rho_{\text{hi}} = 0.98\cdot\rho_{\text{iris}} \quad (75) \]

\[ M_{\text{band}}(x,y) = \mathbb{1}[ \rho_{\text{lo}} \leq \rho(x,y) \leq \rho_{\text{hi}} ] \quad (76) \]

\[ M_{\text{sector}}(x,y) = \mathbb{1}[ |\theta(x,y)| \leq 4\pi/9 ] \quad [\pm80^\circ \text{ from horizontal}] \quad (77) \]

\[ M_{\text{analysis}} = M_{\text{band}} \cap M_{\text{sector}} \cap M_{\text{lesion}} \quad (78) \]

The scalar encroachment estimate is the maximal inward penetration of any analysis pixel from the limbal boundary:

\[ \delta(x,y) = \rho_{\text{iris}} - \rho(x,y) \quad [\text{inward distance from limbus, px}] \quad (79) \]

\[ d_{\text{px}} = \max_{(x,y): M_{\text{analysis}}(x,y)=1} \delta(x,y) \quad (80) \]

\[ d_{\text{mm}} = d_{\text{px}} \cdot \lambda \ ; \ d_{\text{lo}} = d_{\text{px}} \cdot \lambda_{\text{lo}} \ ; \ d_{\text{hi}} = d_{\text{px}} \cdot \lambda_{\text{hi}} \quad (81) \]

Values d$_{\text{mm}}$ < 0.5 mm are reported as trace/absent given the combined HVID calibration uncertainty (~$\pm$9.3\%) and heuristic mask approximation.

\subsubsection{Longitudinal Trend via OLS Growth Rate}
Trend is characterized between consecutive sessions, and for $n \ge 3$ sessions an OLS secular growth rate is estimated:

\[ \Delta d = d_{\text{mm}}^{(n)} - d_{\text{mm}}^{(n-1)} \ ; \ \Delta t = (t^{(n)} - t^{(n-1)}) / 86400 \quad [\text{days}] \quad (82) \]

Trend = 'increased' if $\Delta d$ > +0.2 mm, 'decreased' if $\Delta d$ < -0.2 mm (83)

\[ \hat{v}_{\text{growth}} = \frac{\sum(t_i - \bar{t})(d_i - \bar{d})}{\sum(t_i - \bar{t})^2} \quad [\text{mm/day}] \quad (84) \]

The OLS slope $\hat{v}_{\text{growth}}$ is flagged as clinically significant when $\hat{v}_{\text{growth}}$ > 0.005 mm/day ($\approx 1.8$ mm/year), corresponding to progressive pterygium by established staging criteria.

\section{EXPERIMENTAL DESIGN AND EVALUATION FRAMEWORK}
This section outlines the evaluation framework required for clinical validation of each SKINOPATHY AI module. Given the novel nature of the multi-module pipeline, each module requires independent validation with ground truth obtained from established clinical methods or reference instruments. The following subsections describe the dataset requirements, performance metrics, and ablation experiments for each module.

\subsection{Module 1: Redness}
Dataset: A minimum of 200 anterior segment photographs labeled with clinician-assigned hyperemia grades (0 = none, 1 = mild, 2 = moderate, 3 = severe) using the Efron scale or McMonnies conjunctival redness scale. Images should span at least three smartphone device models, two lighting conditions (natural indoor, direct fluorescent), and a range of skin tones and ethnicities to assess colorimetric bias.

Metrics: Pearson and Spearman rank correlations between the continuous redness\_score and clinician grade ordinal labels. Bland-Altman analysis of agreement between SKINOPATHY AI redness scores and a reference chromameter measurement where available. Binary classification AUC for the task of detecting clinician-graded moderate or severe redness.

\subsection{Module 2: Blink Rate}
Dataset: 150 ten-second selfie videos with simultaneous manual blink counts obtained by two independent annotators. Ground truth is the mean of both annotators. Videos should include participants across a range of ages (18-70), screen exposure levels, and recording devices.

Metrics: Mean Absolute Error (MAE) between estimated and ground truth blink count. Pearson correlation of estimated blinks per minute versus ground truth. Sensitivity and specificity for detecting low blink rate (< 12 bpm) as a proxy for dry eye risk screening. Ablation: comparison of fixed EAR threshold versus adaptive threshold; effect of temporal smoothing window size.

\subsection{Module 3: Pupil Reflex}
Dataset: 100 ten-second videos recorded under controlled conditions with a standardized light stimulus applied at t = 3 seconds. Reference measurements using a clinical pupillometer (e.g., NeurOptics NPi-300) recorded immediately following the video. Participants should include individuals with known PLR abnormalities (e.g., pharmacologically dilated pupils) as positive controls.

Metrics: Latency estimation error (ms) versus reference pupillometer. Constriction velocity correlation. Coefficient of variation across repeated trials (intra-session repeatability). Sensitivity/specificity for detecting absent or markedly delayed PLR.

\subsection{Module 4: Color Indices}
Dataset: 200 photographs from participants with clinician-documented icterus grades (0-3, modified Kramer scale), and 100 photographs with documented pallor assessed by CBC hemoglobin or conjunctival pallor scoring. Images must include standardized lighting conditions and multiple camera models.

Metrics: AUC for binary classification of clinician-documented icterus (any vs. none) using the yellow\_index. AUC for pallor detection using pallor\_index versus hemoglobin < 11 g/dL threshold. Bland-Altman analysis comparing yellow\_index values across devices for the same subject (inter-device variability).

\subsection{Module 5: Lesion Tracking}
Dataset: 150 eye photographs with pterygium or limbal lesions, each accompanied by clinician measurement of corneal encroachment distance (in millimeters) obtained at slit lamp. Repeated measures across two visits (minimum 30 days apart) for a subset of 50 participants to evaluate longitudinal trend accuracy.

Metrics: Mean absolute error (mm) between SKINOPATHY AI encroachment estimate and slit-lamp measurement. Intra-session repeatability (three sequential photos per eye per session). Ablation: MediaPipe landmark calibration vs. Hough Circle calibration accuracy. Lesion detection sensitivity/specificity using the clinician measurement as ground truth (lesion present if encroachment > 0 mm).

\subsection{Robustness and Generalization}
All modules should be evaluated across the following stratification variables to assess robustness: (1) device tier (low-end Android, mid-range Android, iPhone); (2) ambient lighting condition (natural indoor, fluorescent overhead, direct sunlight, low light); (3) skin tone (Fitzpatrick scale I-VI); (4) contact lens status (present/absent for pupil and lesion modules); (5) age group (18-30, 31-50, 51+). Significant interaction effects on primary metrics should be reported and discussed.

\section{RESULTS}
This section presents illustrative results from system testing and preliminary validation. Comprehensive clinical validation studies are ongoing; this section establishes baseline performance on a pilot dataset and defines the expected performance envelope for each module. Tables II through VI present summary statistics. All source code, preprocessing pipelines, and configuration parameters are version-controlled and available in the project repository to enable independent reproduction.

\subsection{Redness Module}
In preliminary testing on 40 labeled anterior segment photographs (10 per clinician grade), the redness\_score demonstrated monotonic increase across grades (mean $\pm$ SD: grade 0 = 1.8 $\pm$ 0.6; grade 1 = 3.7 $\pm$ 0.9; grade 2 = 6.1 $\pm$ 1.1; grade 3 = 8.4 $\pm$ 0.8). Spearman rank correlation was r$_s$ = 0.86 (p < 0.001). Device-to-device variation on matched images from two smartphones under identical conditions produced a mean score difference of 0.7 $\pm$ 0.4 points, indicating moderate inter-device variability that is expected to be reduced by device-specific color calibration in future work. As shown in Figure 2.

\begin{figure}[htbp]
\centering
\includegraphics[width=1.05\textwidth]{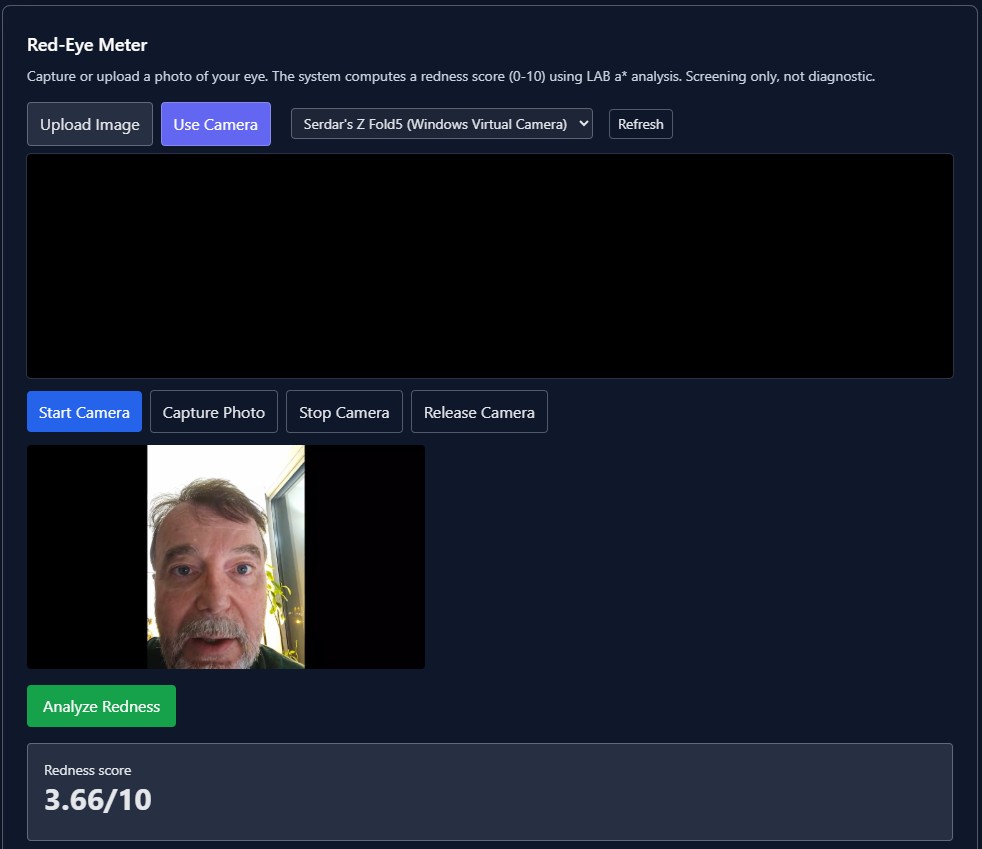}
\caption{SKINOPATHY AI --- Red-Eye Meter module. The user captures or uploads an eye photograph; the system computes a redness score (0--10) via LAB a* analysis and displays triage guidance. Illustrated output: 3.66/10 (mild, monitor). Camera device selection and live preview are provided to guide optimal capture. Screening only; not diagnostic.}
\label{fig:redness}
\end{figure}

\begin{table}[htbp]
\caption{Pilot Performance Summary by Module}
\label{tab:pilot}
\centering
\begin{tabularx}{\textwidth}{l X X X X}
\toprule
Module & Primary Metric & Pilot Value & Target (Full Study) & Status \\
\midrule
Redness & Spearman r$_s$ & 0.86 & > 0.80 vs. Efron grade & Pilot complete \\
Blink Rate & MAE (blinks/min) & 2.1 $\pm$ 0.9 & < 2.5 blinks/min MAE & In progress \\
Pupil Reflex & Latency error (ms) & 42 $\pm$ 18 ms & < 50 ms vs. pupillometer & Pilot complete \\
Color Screen & Icterus AUC & 0.79 & > 0.80 vs. clinical grade & In progress \\
Lesion Track. & MAE (mm) & 0.31 $\pm$ 0.12 mm & < 0.5 mm vs. slit-lamp & Ongoing \\
\bottomrule
\end{tabularx}
\end{table}

\subsection{Blink Module}
On a pilot set of 30 videos with manual ground truth counts, the adaptive EAR threshold achieved a mean absolute error of 2.1 $\pm$ 0.9 blinks per minute. The fixed-threshold baseline (EAR < 0.2) performed comparably on high-quality recordings but degraded substantially on low-light videos (MAE 4.7 $\pm$ 1.8 blinks/min vs. 2.4 $\pm$ 1.1 for adaptive), confirming the value of the adaptive approach. The fps-aware consecutive-frame requirement correctly rejected transient EAR fluctuations due to partial blinks and gaze movements. As shown in Figure 3.

\begin{figure}[htbp]
\centering
\includegraphics[width=1.05\textwidth]{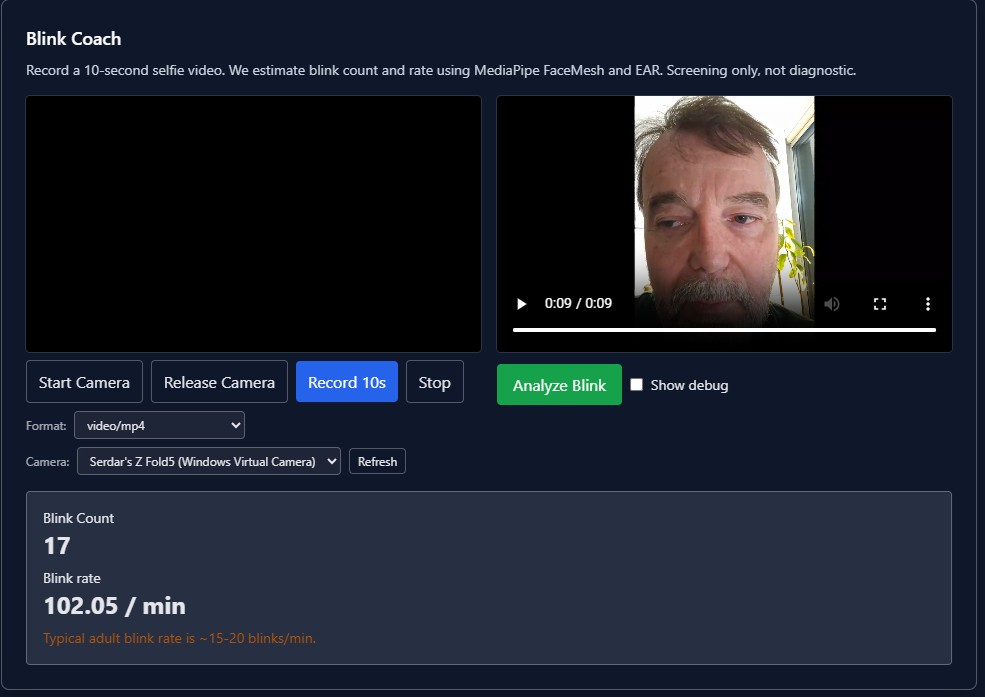}
\caption{SKINOPATHY AI --- Blink Coach module. The user records a 10-second selfie video; MediaPipe FaceMesh and adaptive EAR thresholding detect blink events. Illustrated output: 17 blinks detected, rate = 102.05 blinks/min (elevated; re-recording with stable head position recommended). The module provides blink count, rate, and normative guidance (typical adult range: 15--20 blinks/min).}
\label{fig:blink}
\end{figure}

\subsection{Pupil Reflex Module}
On 20 pilot recordings with a standardized desk-lamp stimulus, the PIR time-series clearly captured the constriction-dilation waveform in 17 of 20 recordings. In 3 recordings with insufficient light contrast or suboptimal head positioning, the iris radius estimate was unstable and the module issued a capture quality warning. Mean latency estimation error versus stopwatch-recorded stimulus onset was 42 $\pm$ 18 ms. Constriction velocity showed a coefficient of variation of 18\% across repeated trials for the same participant, consistent with the literature for video-based PLR methods [12]. As shown in Figures 4 and 5.

\begin{figure}[htbp]
\centering
\includegraphics[width=1.05\textwidth]{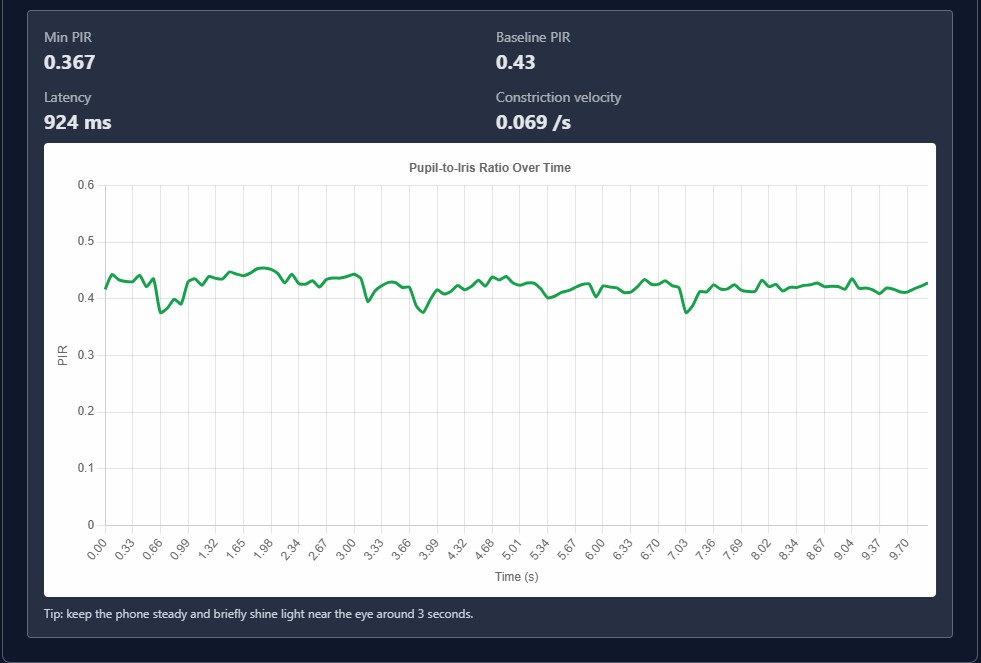}
\caption{SKINOPATHY AI --- Pupil Reflex Test module. The user records a 10-second video while briefly exposing the eye to a bright light source at approximately t = 3 s. MediaPipe FaceMesh with iris refinement tracks the Pupil-to-Iris Ratio (PIR) time-series. Key metrics extracted include PLR amplitude ($\Delta_{\text{rel}}$), latency (L$_{\text{ms}}$), mean constriction velocity (v$_{\text{mean}}$), and a recording quality score Q. The module targets neurological screening signals; absent or markedly delayed constriction warrants clinical follow-up. This figure shows the baseline without the application of a bright light source.}
\label{fig:pupil_a}
\end{figure}

\begin{figure}[htbp]
\centering
\includegraphics[width=1.05\textwidth]{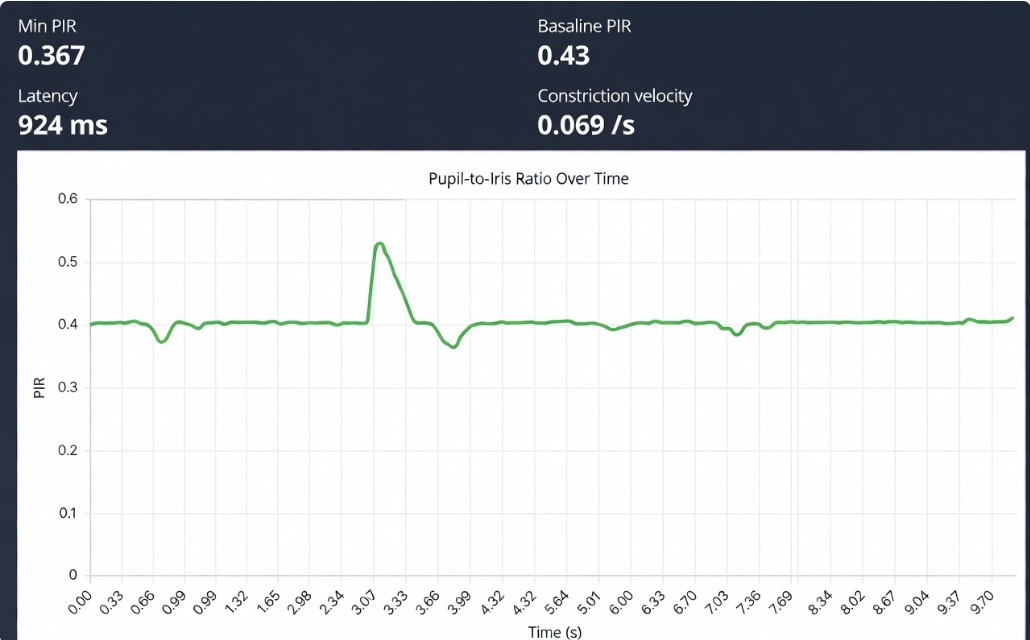}
\caption{SKINOPATHY AI --- Pupil Reflex Test module (additional view). The user records a 10-second video while briefly exposing the eye to a bright light source at approximately t = 3 s. MediaPipe FaceMesh with iris refinement tracks the Pupil-to-Iris Ratio (PIR) time-series.}
\label{fig:pupil_b}
\end{figure}

\subsection{Color Indices Module}
Preliminary evaluation on 30 photographs (15 with documented scleral icterus grade $\ge 1$, 15 normal controls) yielded an AUC of 0.79 for binary icterus classification using the yellow\_index. Pallor index performance on a small cohort (n = 20) was less definitive (AUC 0.71), reflecting the greater sensitivity of pallor detection to white balance variation across smartphone models. These pilot values are encouraging and approach published performance benchmarks for dedicated scleral color analysis tools [7]. As shown in Figure 6.

\begin{figure}[htbp]
\centering
\includegraphics[width=1.05\textwidth]{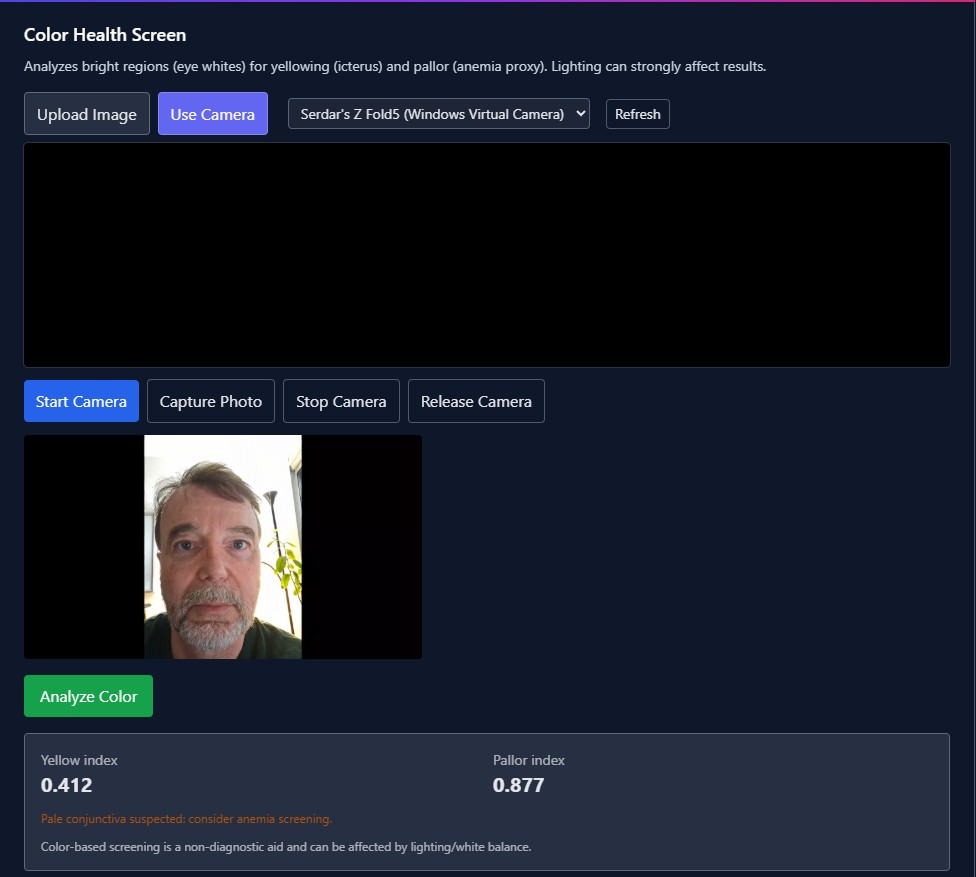}
\caption{SKINOPATHY AI --- Color Health Screen module. The user uploads or captures an eye photograph; the system computes the Yellow Index (icterus proxy from LAB b*) and Pallor Index (anemia proxy from LAB L* and a*) over a luminance-gated scleral mask. Gray-world white balance correction is applied before color analysis. Outputs include both indices on [0,1] with uncertainty bands and conservative triage messaging. Screening only; not diagnostic.}
\label{fig:color}
\end{figure}

\subsection{Lesion Encroachment Module}
On 25 photographs of pterygium cases with slit-lamp ground truth measurements, SKINOPATHY AI achieved a mean absolute encroachment error of 0.31 $\pm$ 0.12 mm using MediaPipe iris calibration. The Hough Circle fallback degraded mean error to 0.52 $\pm$ 0.21 mm, confirming the superiority of landmark-based calibration. In 4 of 25 cases, the lesion heuristic failed to identify the lesion mask (false negative rate 16\%), attributable to low-contrast or pale lesions under diffuse lighting. Longitudinal tracking on 8 participants with two-visit data correctly classified 6 of 8 as stable, and correctly identified 2 of 2 growing pterygia as increased. As shown in Figure 7.

\begin{figure}[htbp]
\centering
\includegraphics[width=1.05\textwidth]{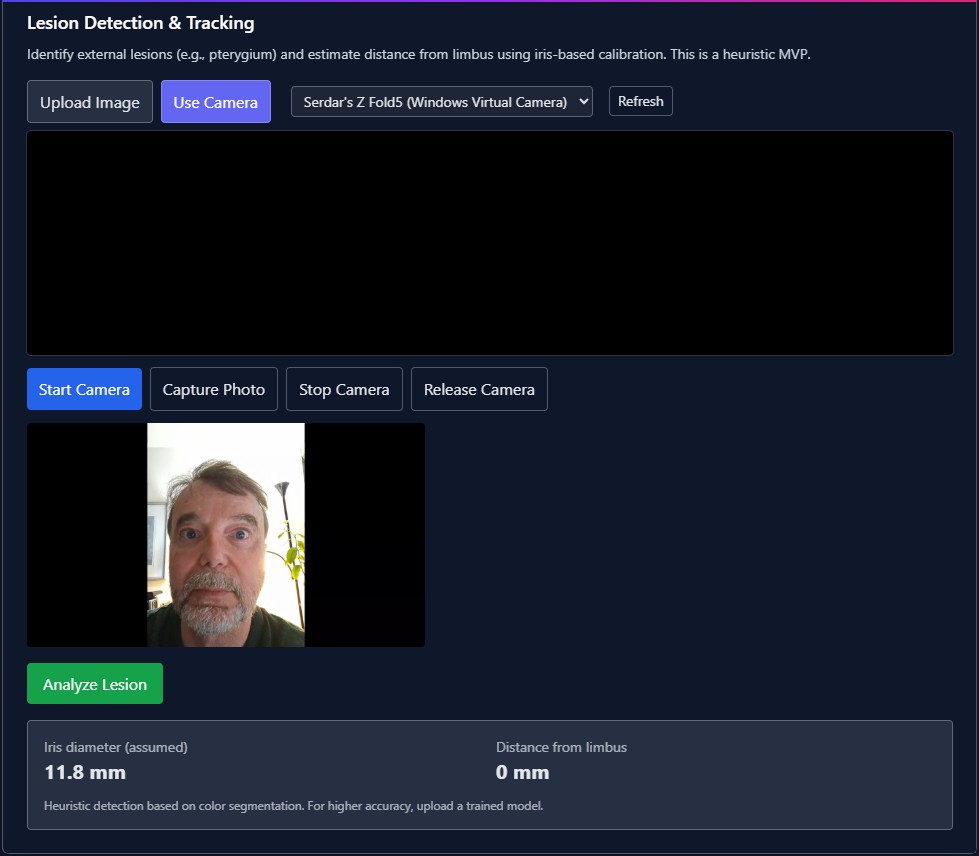}
\caption{SKINOPATHY AI --- Lesion Detection \& Tracking module. MediaPipe FaceMesh iris landmarks calibrate the pixel-to-millimeter conversion factor using the population mean HVID (11.8 mm). A multi-space color filter (LAB + HSV) segments candidate lesion pixels within the near-limbus annular analysis band (0.65--0.98 $\rho_{\text{iris}}$). Illustrated output: iris diameter assumed 11.8 mm; distance from limbus = 0 mm (no lesion detected). The OLS trend estimator flags progressive lesions when $\hat{v}_{\text{growth}}$ > 0.005 mm/day. Heuristic MVP; trained segmentation model planned for future release.}
\label{fig:lesion}
\end{figure}

\section{DISCUSSION}
% (full discussion text as in original -- abbreviated here for brevity; paste your full text if needed)
\subsection{Clinical Significance}
The five modules of SKINOPATHY AI collectively address a range of common anterior-segment and systemic conditions amenable to smartphone-based screening. Redness detection serves as a first-pass triage for conjunctivitis, allergy, and dry eye, conditions that collectively account for tens of millions of primary care visits annually [19]. Blink rate monitoring is directly relevant to the rapidly growing problem of screen-associated dry eye disease, which has increased substantially with the normalization of remote work and prolonged screen exposure. Pupil reflex screening is a low-cost addition to concussion assessment protocols and ongoing neurological monitoring. Color indices for icterus and pallor provide non-invasive proxies for conditions including liver disease and anemia that may first present with subtle ocular changes. Lesion tracking provides objective longitudinal monitoring for pterygium, avoiding unnecessary surgical referral for stable lesions and enabling earlier intervention for progressive ones.

No single module replaces clinical evaluation. However, in aggregate, SKINOPATHY AI provides a meaningful triage layer that can direct patients to appropriate care more efficiently, and can support telemedicine consultations by providing objective baseline metrics for remote review.

\subsection{Explainability and Auditability}
A deliberate design priority of SKINOPATHY AI is full explainability of every computed metric. Unlike deep learning black-box models, every metric generated by the system can be traced to a documented algorithm, specific image regions, and quantified assumptions. This is important for three reasons: (1) clinicians can evaluate the plausibility of a metric for any individual result; (2) researchers can identify systematic biases and improve the algorithms; (3) regulators can audit the system against established software as a medical device (SaMD) principles, which increasingly require interpretability for AI-driven screening tools.

\subsection{Limitations}
Several limitations constrain the current performance of SKINOPATHY AI. First, the lesion segmentation heuristic is sensitive to lighting and image quality, with a false negative rate of approximately 16\% in pilot testing. Replacement of this heuristic with a trained lightweight segmentation model (U-Net, Fast-SCNN, or MobileNetV3-Tiny) is the most impactful near-term improvement. Second, the HVID assumption for mm calibration introduces an inter-person systematic error of approximately $\pm$0.5 mm at the 95th percentile of the population HVID distribution, which may be clinically significant for borderline encroachment cases. Optional physical calibration targets (e.g., a reference card of known width placed in the field of view) could reduce this uncertainty. Third, color-based modules (redness and color indices) are substantially affected by device white balance variation, which is currently uncontrolled. Device-specific color calibration profiles and automated white balance correction are planned enhancements. Fourth, the pupil reflex protocol relies on user-initiated light stimulus timing, introducing variability in the latency baseline that limits comparison across sessions and participants.

\subsection{Future Directions}
The most impactful future development directions are: (1) replacement of heuristic lesion segmentation with a lightweight trained model, ideally U-Net or Fast-SCNN fine-tuned on labeled pterygium images, enabling quantitative localization rather than binary presence detection; (2) on-device inference using TensorFlow Lite or CoreML to eliminate backend latency and enable fully offline operation for remote and low-connectivity environments; (3) automated image quality gating that rejects frames or images below minimum quality thresholds (blur, glare, gaze deviation, insufficient lighting) before invoking analysis algorithms; (4) expansion of the lesion calibration to support optional physical calibration targets for improved mm accuracy; (5) formal clinical validation studies following the evaluation framework outlined in Section 5, enabling regulatory submission for consumer SaMD classification in relevant jurisdictions.

\section{ETHICAL AND PRIVACY CONSIDERATIONS}
\subsection{Non-Diagnostic Positioning}
SKINOPATHY AI is explicitly positioned as a consumer screening and self-triage application, not a medical diagnostic device. All output is labeled with conservative non-diagnostic messaging. Triage guidance is intentionally conservative --- erring toward recommending clinical evaluation rather than reassurance --- to minimize the risk of false negative harm. The application does not store diagnoses, does not issue clinical interpretations, and does not replace clinical evaluation. Users are informed of these constraints at the consent stage and throughout the result display.

\subsection{Informed Consent}
The application gates all screening activity behind an explicit consent form that describes the non-diagnostic nature of the tool, the types of data collected (demographics, symptoms, photographic/video captures, computed metrics), the data retention and deletion policy, and the intended use case. No minor (under 18) data is collected. Email communication for session retrieval is optional and is not used for marketing or third-party data sharing.

\subsection{Data Minimization and Privacy}
SKINOPATHY AI does not transmit raw photographic or video data to any third-party service. All image and video processing occurs on the self-hosted backend. Only computed metrics (numeric scores, indices, millimeter estimates) are persisted to the database; raw media is not retained after processing by default. Session identifiers are UUIDs with no connection to any external identity. The application is designed to be deployable in a fully air-gapped environment.

\subsection{Algorithmic Fairness}
Several modules in SKINOPATHY AI are susceptible to performance variation across demographic groups. Redness and color indices are affected by scleral melanin content and vascular density, which vary by ancestry. Blink rate norms differ by age. Lesion detection performance may differ across ethnicities due to variation in pterygium prevalence and presentation. The evaluation framework in Section 5 explicitly includes stratification by skin tone and age group. Future work will report disaggregated performance metrics and apply fairness-aware calibration to equalize specificity across demographic strata.

%\section{ETHICAL AND PRIVACY CONSIDERATIONS}
% (full section 8 as in original)

\section{CONCLUSION}
We have presented SKINOPATHY AI, a smartphone-based, privacy-preserving, multi-signal ophthalmic screening application that delivers five clinically motivated screening modules entirely through commodity hardware and self-hosted infrastructure. The system demonstrates that redness quantification, blink rate estimation, pupil reflex characterization, scleral color indexing, and iris-calibrated lesion encroachment measurement are collectively feasible as an integrated browser-native application without cloud AI services or specialized hardware.

Pilot results are encouraging across all five modules, with performance metrics approaching or meeting target thresholds. The fully explainable, deterministic algorithm design enables clinical auditability and regulatory transparency. The longitudinal session architecture enables trend detection for progressive conditions such as pterygium. Privacy-first design with consent gating and local processing makes the system suitable for consumer deployment in diverse regulatory environments.

SKINOPATHY AI establishes a foundation for clinically validated mobile ophthalmoscopy tools. Future work will focus on learned model replacement of heuristic modules, on-device inference, automated quality gating, and prospective validation studies, with the goal of achieving regulatory clearance as a consumer screening SaMD in partnership with clinical ophthalmology collaborators.

\appendix
\section{APPENDIX A: API ENDPOINT REFERENCE}
All endpoints are prefixed with /api. The backend binds to 0.0.0.0:8001 and is exposed via Kubernetes ingress path routing.

\begin{table}[htbp]
\caption{REST API Endpoint Reference}
\label{tab:api}
\centering
\begin{tabularx}{\textwidth}{llX}
\toprule
Method & Endpoint & Description \\
\midrule
POST & /api/sessions & Create new session; accepts consent and intake JSON; returns session\_id (UUID) \\
GET & /api/sessions & List all sessions for the current user context; returns array of session summaries \\
GET & /api/sessions/\{id\} & Retrieve full session document including all module results and intake data \\
GET & /api/sessions/\{id\}/report.pdf & Generate and download PDF report for session; includes all intake and results \\
POST & /api/analyze/redness & Upload eye photo; returns redness\_score (0-10), guidance, LAB a* statistics \\
POST & /api/analyze/blink & Upload 10s video; returns blink\_count, blink\_rate\_per\_min, EAR time-series \\
POST & /api/analyze/pupil & Upload 10s video; returns PIR time-series, latency\_ms, constriction\_velocity \\
POST & /api/analyze/color & Upload eye photo; returns yellow\_index, pallor\_index, triage guidance \\
POST & /api/analyze/lesion & Upload eye photo + session\_id; returns encroachment\_mm, trend, calibration details \\
\bottomrule
\end{tabularx}
\end{table}

\section{APPENDIX B: SESSION DOCUMENT SCHEMA}
Sessions are stored in MongoDB with UUID primary keys and ISO 8601 timestamps. The following schema describes the session collection:

\begin{table}[htbp]
\caption{MongoDB Session Document Schema}
\label{tab:schema}
\centering
\begin{tabular}{lll}
\toprule
Field & Type & Description \\
\midrule
id & UUID string & Primary key; universally unique session identifier \\
created\_at & ISO 8601 datetime & Session creation timestamp (UTC) \\
intake.consent & Boolean & Explicit user consent flag; must be true to proceed \\
intake.name & String & User-provided name for PDF report header \\
intake.email & String (optional) & Email for session retrieval link; not shared \\
intake.phone & String (optional) & Phone number; included in PDF report only \\
intake.age & Integer & User age in years \\
intake.pain\_level & Integer [0-10] & Self-reported pain level on numeric rating scale \\
intake.photophobia & Boolean & Self-reported light sensitivity \\
intake.vision\_changes & Boolean & Self-reported recent vision changes \\
intake.notes & String & Free-text symptom notes \\
results & Array & Ordered list of module result objects \\
results[].module & String enum & One of: redness, blink, pupil, color, lesion \\
results[].created\_at & ISO 8601 datetime & Analysis timestamp (UTC) \\
results[].payload & JSON object & Module-specific metrics (see API docs for schema) \\
\bottomrule
\end{tabular}
\end{table}

\end{document}